\documentclass[10pt,twocolumn,letterpaper]{article}

\usepackage{iccv}              

\usepackage{makecell}
\definecolor{iccvblue}{rgb}{0.21,0.49,0.74}
\usepackage[pagebackref,breaklinks,colorlinks,allcolors=iccvblue]{hyperref}

\def\paperID{22} 
\def\confName{ICCV}
\def\confYear{2025}

\title{Robust Vision-Language Models via Tensor Decomposition: A Defense Against Adversarial Attacks}

\author{Het Patel \quad
        Muzammil Allie \quad
        Qian Zhang \quad
        Jia Chen \quad
        Evangelos E. Papalexakis\\
        University of California, Riverside\\
        {\tt\small \{hpate061, malli007, qzhang, jiac, epapalex\}@ucr.edu}
}

\begin{document}
\maketitle

\begin{abstract}
Vision language models (VLMs) excel in multimodal understanding but are prone to adversarial attacks. Existing defenses often demand costly retraining or significant architecture changes. We introduce a lightweight defense using tensor decomposition suitable for any pre-trained VLM, requiring no retraining. By decomposing and reconstructing vision encoder representations, it filters adversarial noise while preserving meaning. Experiments with CLIP on COCO and Flickr30K show improved robustness. On Flickr30K, it restores 12.3\% performance lost to attacks, raising Recall@1 accuracy from 7.5\% to 19.8\%. On COCO, it recovers 8.1\% performance, improving accuracy from 3.8\% to 11.9\%. Analysis shows Tensor Train decomposition with low rank (8-32) and low residual strength ($\alpha=0.1-0.2$) is optimal. This method is a practical, plug-and-play solution with minimal overhead for existing VLMs.
\end{abstract}

\section{Introduction}
Vision language models (VLMs) such as CLIP\cite{radford2021learningtransferablevisualmodels} have revolutionized multimodal understanding by learning joint embeddings of images and text, enabling applications like image retrieval\cite{lahajal2024enhancingimageretrieval}, zero-shot classification\cite{qian2024onlinezeroshotclassificationclip}, and visual question answering\cite{ye2023videoquestionansweringusing}. However, VLMs remain vulnerable to adversarial attacks\cite{mao2023understandingzeroshotadversarialrobustness, wang2025doublevisualdefenseadversarial}, where imperceptible perturbations dramatically degrade performance. Ensuring robustness is essential as these models deploy in critical applications.

Traditional defenses like adversarial training\cite{madry2019deeplearningmodelsresistant}, input preprocessing\cite{xie2017adversarialexamplessemanticsegmentation}, and architectural modifications\cite{papernot2016distillationdefenseadversarialperturbations} have significant limitations: adversarial training requires expensive retraining on large datasets, while architectural changes demand substantial redevelopment effort. For large pre-trained VLMs, these solutions are often impractical due to computational constraints or unavailable training data.

We propose a novel tensor decomposition defense applicable to any pretrained VLM without retraining or modification. Leveraging the fact that adversarial perturbations introduce high-frequency noise in feature representations\cite{lorenz2024detectingautoattackperturbationsfrequency}, we decompose and reconstruct intermediate representations using low-rank tensor approximations to filter perturbations while preserving semantic content.

Our contributions include: (1) a lightweight tensor decomposition defense requiring no retraining, (2) comprehensive parameter analysis, (3) empirical robustness improvements on COCO and Flickr30K, and (4) publicly available code at \href{https://github.com/HettyPatel/TensorDefenseVLM}{GitHub Link}.

\section{Related Work}

Defensive Tensorization \cite{bulat2021defensivetensorization} applies Tucker decomposition to express CNN weight tensors in lower-dimensional latent spaces, introducing tensor dropout to stochastically reduce tensor ranks and prevent adversarial exploitation of fixed-weight structures.

Input-level defenses include SmoothVLM\cite{robey2024smoothllmdefendinglargelanguage}, which uses random pixel perturbations with majority voting(achieving 95\% attack reduction), and TensorSHIELD\cite{entezari2020tensorshieldtensorbaseddefenseadversarial}, which filters adversarial noise via low-rank tensor decompositions. FARE\cite{schlarmann2024robustclipunsupervisedadversarial} utilizes unsupervised alignment between perturbed and original images for consistent outputs withouit labeled data.



\begin{figure}
\centering
\includegraphics[width=0.32\linewidth]{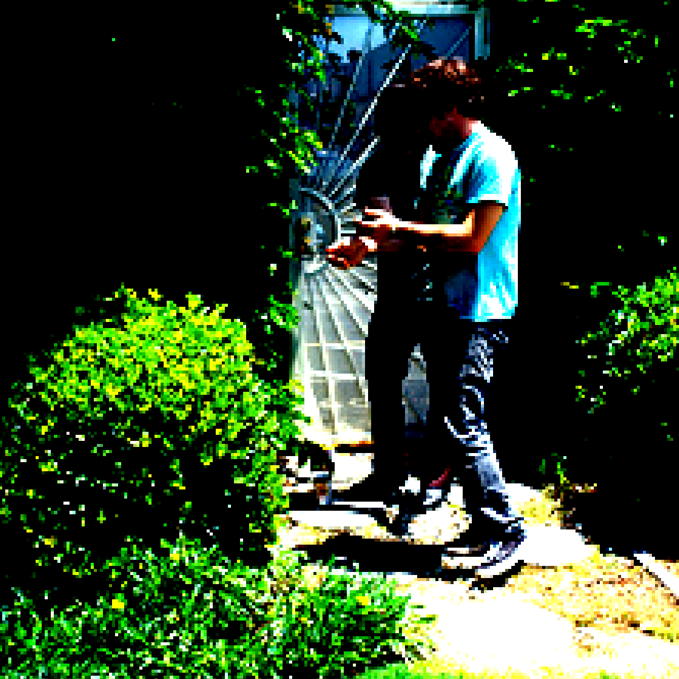}
\includegraphics[width=0.32\linewidth]{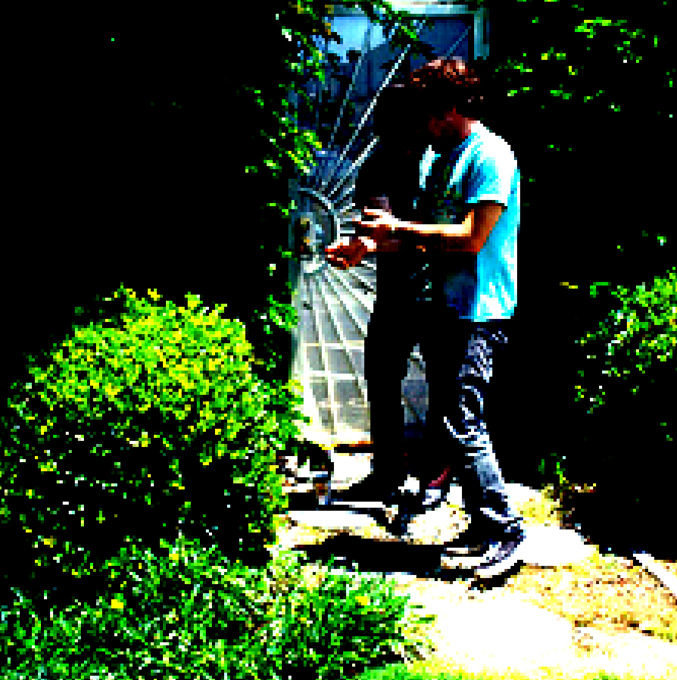}
\includegraphics[width=0.32\linewidth]{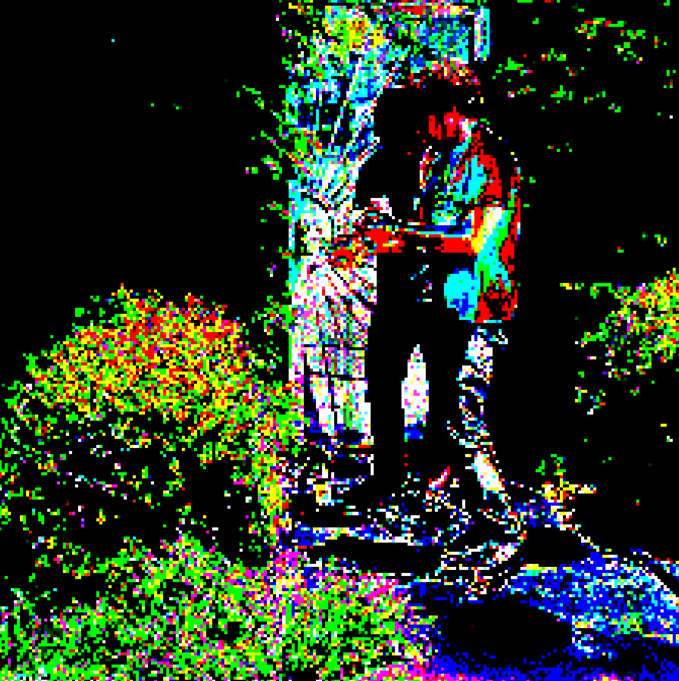}\\
\small (a) Original \hfill (b) Adversarial \hfill (c) Perturbation
\caption{Adversarial attack visualization: original image, adversarial version ($\epsilon=0.0314$), and scaled perturbation pattern.}
\label{fig:adversarial_example}
\end{figure}

Double Visual Defense\cite{wang2025doublevisualdefenseadversarial} improves the robustness using a two-stage adversarial training process: extensive pretraining to boost zero-shot capabilities and attack resistance, followed by fine-tuning on downstream tasks with adversarially perturbed inputs. 

Current approaches have notable limitations: Defensive Tensorization factorizes weight matrices without considering input variations, SmoothVLM requires multiple forward passes per query, TensorSHIELD lacks dynamic adaptation, and FARE and Double Visual Defense demand substantial computational resources and frequent dataset updates against evolving attacks. 



\section{Methodology}
We present a defense mechanism against adversarial attacks on a vision-language model utilizing tensor decomposition techniques\cite{8884203, bacciu2020tensordecompositionsdeeplearning, comon2009tensordecompositionsstateart}, including CP/PARAFAC\cite{Harshman1970FoundationsOT}, Tucker\cite{Tucker_1966}, and Tensor-Train Decomposition\cite{doi:10.1137/090752286}. Our approach works by applying low-rank tensor decompositions to the internal representation of the CLIP model.

\subsection{Problem Formulation}

We consider a pre-trained VLM that maps images and text to a joint
embedding space. For an image-text pair $(I, T)$, the model produces
embeddings $f_I(I)$ and $f_T(T)$. During inference, the similarity 
between these embeddings is computed as:
\begin{equation}
s(I, T) = \langle f_I(I), f_T(T) \rangle
\end{equation}
where $\langle \cdot, \cdot \rangle$ denotes the cosine similarity.

An adversarial attack aims to find a perturbed image $I_{adv}$ such that:
\begin{equation}
I_{adv} = I + \delta, \text{ where } \|\delta\|_{\infty} \leq \epsilon
\end{equation}

The perturbation $\delta$ is designed to minimize the similarity between 
the perturbed image and its corresponding text description:
\begin{equation}
\min_{\delta} s(I + \delta, T) \text{ subject to } \|\delta\|_{\infty} \leq \epsilon
\end{equation}

Our goal is to defend against such attacks by restoring the correct
cross-modal alignment in the presence of adversarial perturbations.

\subsection{Adversarial Attack Framework}
We use a white-box approach known as Projected Gradient Descent (PGD) specifically on CLIP models. Given an image-text $(x, t)$ pairs, the attack minimizes the cosine similarity between their embeddings: 

$$L_{adv}(x_{adv}, t) = -\frac{f_I(x_{adv}) \cdot f_T(t)}{||f_I(x_{adv})|| \cdot ||f_T(t)||}$$

where $f_I$ and $f_T$ are the image and text encoders. The adversarial example $x_{adv}$ is generated through iterative updates:

$$x_{adv}^{i+1} = \Pi_{\epsilon}(x_{adv}^{i} + \alpha \cdot \text{sign}(\nabla_x L_{adv}(x_{adv}^{i}, t)))$$

with perturbation budget $\epsilon = 8/255$, step size $\alpha = 6/255$, and $\Pi_{\epsilon}$ projected onto the $\ell_{\infty}$ ball around the original image. The $\ell_{\infty}$-ball constraint ensures that no pixel in the adversarial image differs from the corresponding pixel in the original image by more than $\epsilon$, maintaining perceptual similarity while allowing effective attacks (Fig. \ref{fig:adversarial_example}).

\subsection{Tensor Decomposition Defense}

Our defense mechanism simplifies CLIP vision encoder representations through three components:

\begin{enumerate}
    \item \textbf{Forward Hook Mechanism:} Intercepts tensors at specific layers (final\_norm, attention, MLP outputs).
    
    \item \textbf{Tensor Decomposition:} Applies one of three methods to tensor $T$:
    \begin{itemize}
        \item \textbf{CP:} $T \approx \sum_{r=1}^{R} a_r \circ b_r \circ c_r$ (rank-1 tensor sums)
        \item \textbf{Tucker:} $T \approx \mathcal{G} \times_1 A \times_2 B \times_3 C$ (core tensor + factors)
        \item \textbf{TT:} $T[i_1, \ldots, i_d] \approx G_1[i_1] \cdots G_d[i_d]$ (3D core products)
    \end{itemize}
    
    \item \textbf{Residual Connection:} $T_{\text{final}} = \alpha \cdot T + (1-\alpha) \cdot \hat{T}$, where $\alpha$ controls defense strength.
\end{enumerate}



    
        
        
    

\section{Experimental Setup}
We used CLIP (ViT-B/32) on MS-COCO and Flickr30K datasets, evaluating 1,000 pairs for parameter analysis and 3,000 for comprehensive evaluation. Attacks used PGD with $\ell_\infty$ constraint: $\epsilon = 8/255$, step size $\alpha = 6/255$, and 10 steps, creating imperceptible perturbations that significantly degrade performance (Fig. \ref{fig:adversarial_example}).

Our tensor decomposition defense, implemented with PyTorch and TensorLy\cite{JMLR:v20:18-277}, uses forward hooks with random initialization, 50 iterations, and $10^{-4}$ tolerance. We evaluated five key factors:

\begin{enumerate}[itemsep=1pt]
    \item \textbf{Decomposition Method:} CP, Tucker, and Tensor Train effectiveness
    \item \textbf{Target Layer:} Final norm, attention output, and MLP layers  
    \item \textbf{Rank Parameter:} Values from 16-256 exploring capacity vs. noise filtering
    \item \textbf{Residual Strength:} Weight $\alpha$ (0.1-0.9) balancing original and reconstructed features
    \item \textbf{Multi-layer Configuration:} Single layer up to all 12 transformer layers
\end{enumerate}

\section{Results and Analysis}

\subsection{Parameter Sensitivity Analysis}

\subsubsection{Effect of \texorpdfstring{$\alpha$}{alpha} Value}
\begin{figure}
    \centering
    \includegraphics[width=\columnwidth]{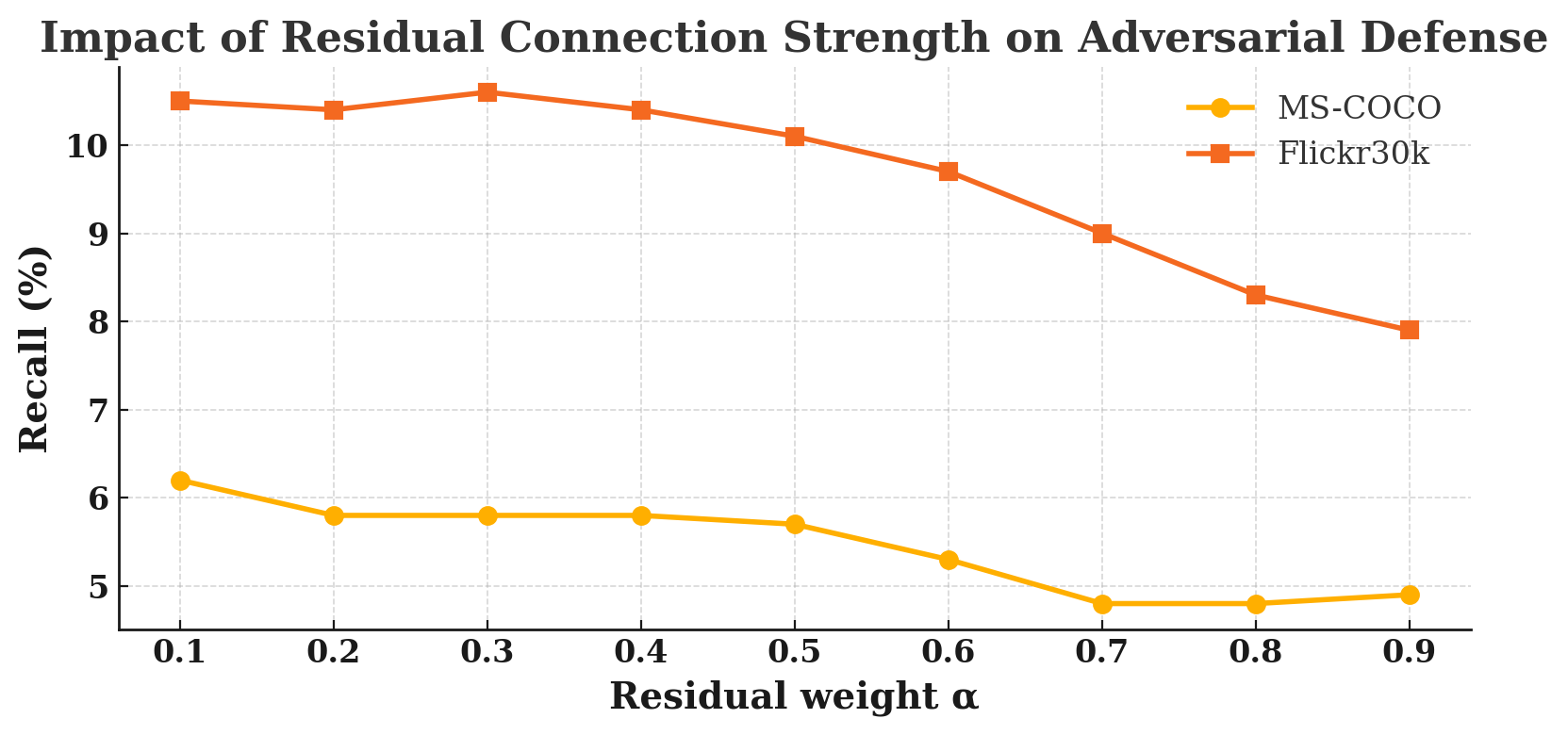}
    \caption{Performance comparison across different $\alpha$ values using Tensor Train decomposition (rank=32) on the final normalization layer. Lower $\alpha$ values provide better defense against adversarial attacks.}
    \label{fig:alpha-sensitivity}
\end{figure}

The effectiveness of tensor decomposition is greatly affected by the residual connection strength parameter $\alpha$, which balances the original and decomposed forms. As seen in \ref{fig:alpha-sensitivity}, there is an inverse relationship between $\alpha$ values and model robustness in the MS-COCO and Flickr30K datasets. Performance peaks at lower $\alpha$ values (0.1-0.3) and declines as $\alpha$ increases. For Flickr30K, the best Recall@1 is 10.6\% at $\alpha=0.3$, and for MS-COCO, it is 6.2\% at $\alpha=0.1$. The downward trend with increasing $\alpha$ supports that adversarial perturbations, as high-frequency components, are effectively filtered by tensor decomposition. 



\subsubsection{Decomposition Rank}
\begin{figure}
    \centering
    \includegraphics[width=\columnwidth]{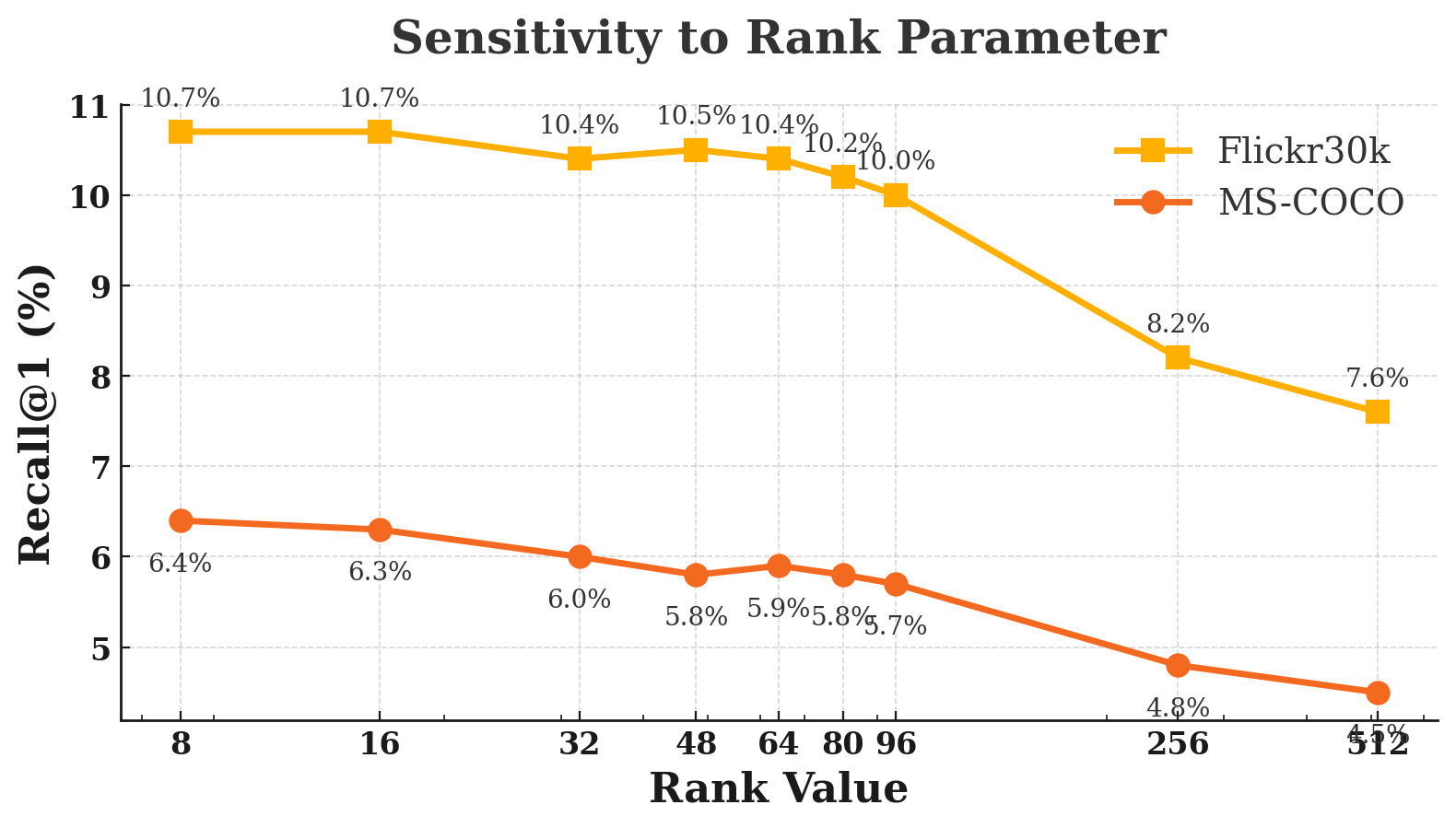}
    \caption{Performance comparison across different rank values using Tensor Train decomposition ($\alpha=0.2$) on the final normalization layer, lower rank values (8-32) provide greater defense against adversarial attacks, while performance degrades significantly at higher ranks.}
    \label{fig:rank-sensitivity}
\end{figure}

The decomposition rank is a critical parameter that determines the representational capacity of the tensor decomposition. Figure \ref{fig:rank-sensitivity} illustrates the effect of varying rank values on adversarial defense performance in the MS-COCO and Flickr30K datasets, with $\alpha$ fixed at 0.2 and applied to the final normalization layer.

Our results reveal an intriguing pattern: lower rank values (8-32) consistently achieve the highest performance on both datasets, with Recall@1 scores of 10.7\% for Flickr30k and 6.3-6.4\% for MS-COCO. Performance remains relatively stable across the 8-96 rank range, with only a modest decline, but drops significantly as rank increases beyond 96. At the highest tested ranks (256-512), performance decreases dramatically to 7.6\% and 4.5\% for Flickr30k and MS-COCO, respectively. 

These findings also validate our assumption that adversarial perturbations manifest themselves primarily as high-frequency components in the feature space. Lower-rank decompositions effectively filter out these perturbations while preserving the essential semantic structure required for cross-modal understanding. As rank increases, the decomposition retains more of the original feature structure, including adversarial noise, reducing the effectiveness of the defense.

\subsubsection{Decomposition Method}

\begin{figure}
    \centering
    \includegraphics[width=\columnwidth]{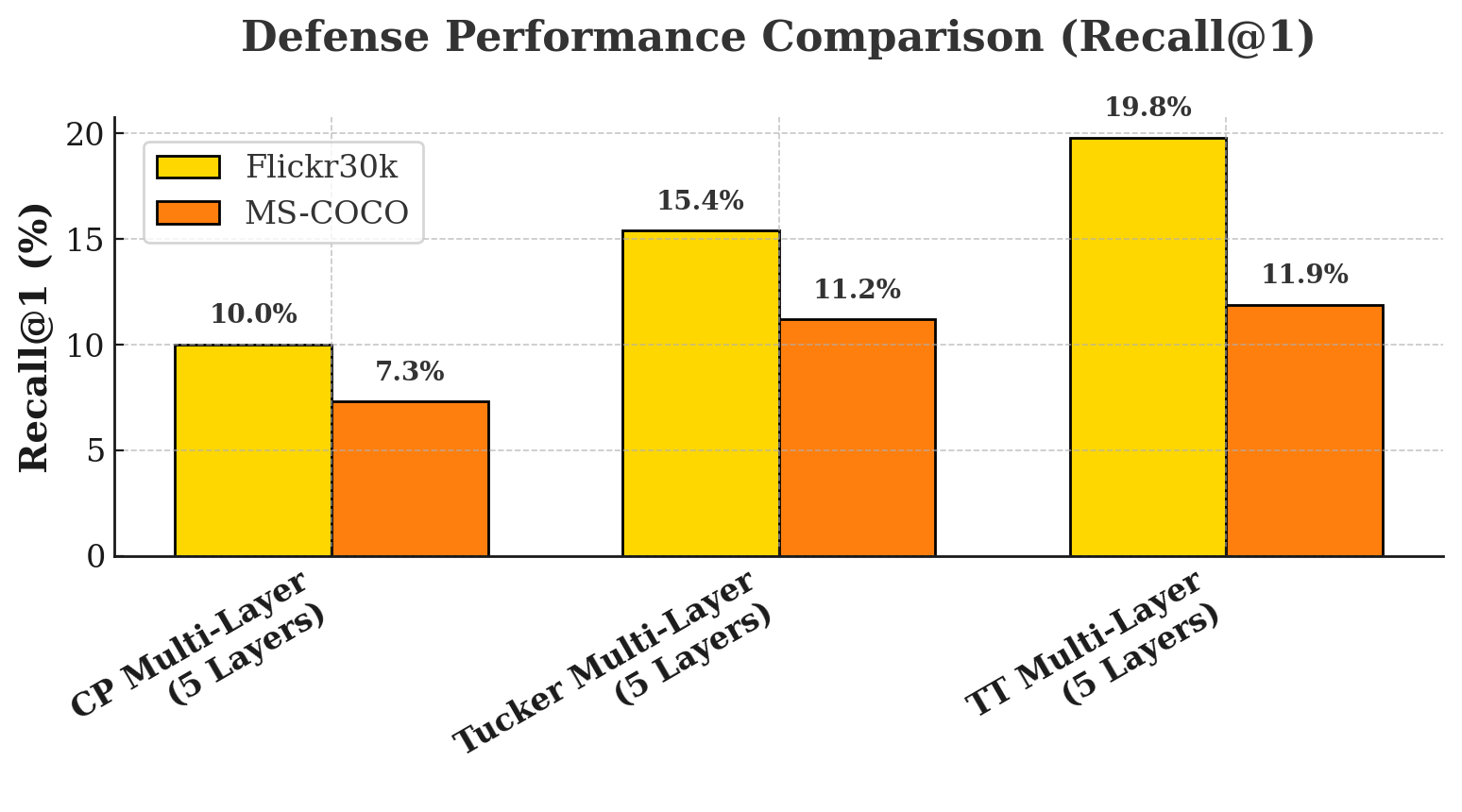}
    \caption{Tensor decomposition methods (CP, Tucker, TT) applied to normalization layers of the last five transformer blocks. All methods substantially improve over adversarial baselines of 3.8\% (MS-COCO) and 7.5\% (Flickr30k), with TT decomposition achieving best performance: 19.8\% (Flickr30k) and 11.9\% (MS-COCO).}

    \label{fig:defense-analysis}
\end{figure}

We evaluated three tensor decomposition methods (CP, Tucker, TT) applied to normalization layers of the last five transformer blocks. Figure \ref{fig:defense-analysis} shows that TT consistently outperforms others: on Flickr30k, TT achieves 19.8\% vs Tucker's 15.4\% and CP's 10.0\%; on MS-COCO, TT reaches 11.9\% vs Tucker's 11.2\% and CP's 7.3\%. This represents 12.3 and 8.1 percentage point recoveries, respectively, from adversarial baselines.

TT's superiority stems from decomposing tensors into three-dimensional core sequences that preserve structural information better than CP's rank-one components, effectively filtering adversarial perturbations while retaining semantic content. Tucker's core tensor structure effectively captures higher-order interactions, whereas CP's simpler rank-one approximation proves insufficient for optimal adversarial filtering.

\subsubsection{Target Layer}
\begin{figure}
    \centering
    \includegraphics[width=\columnwidth]{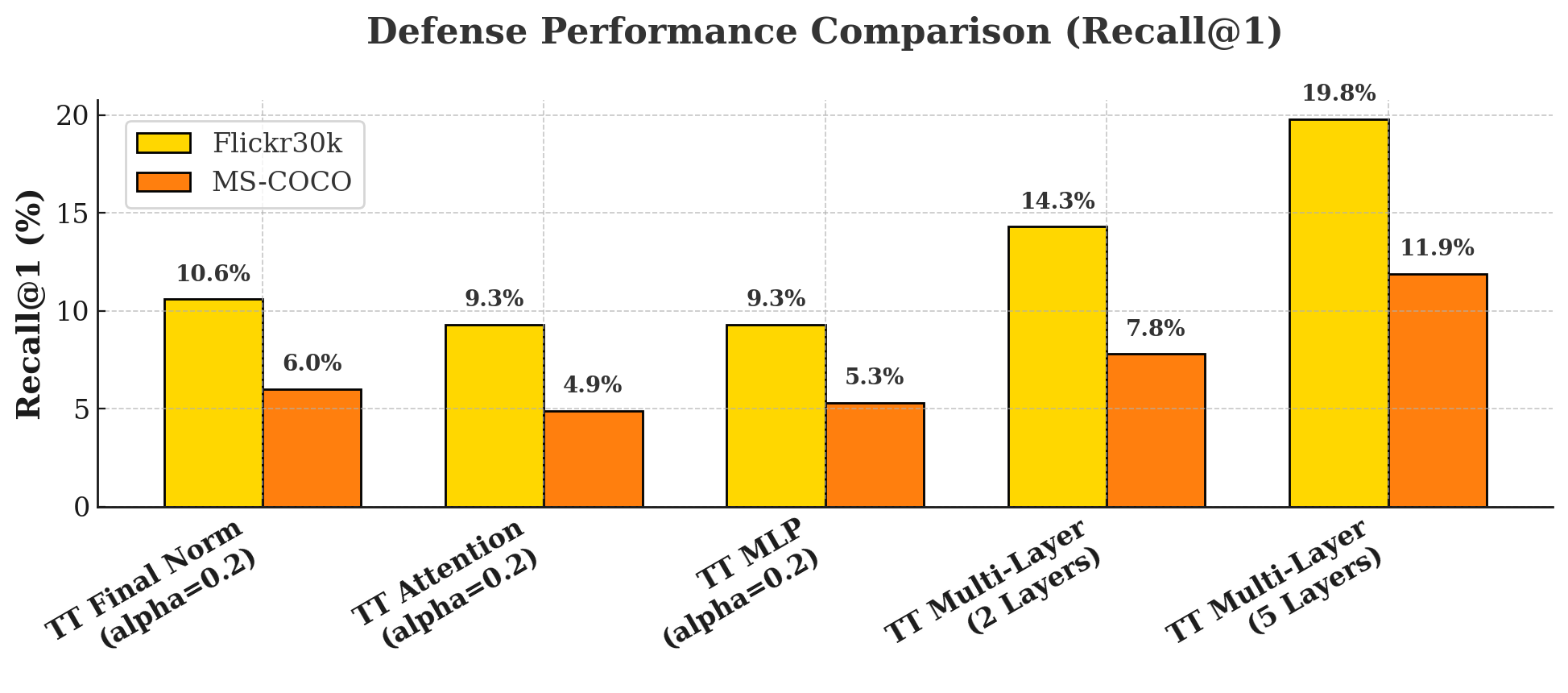}
    \caption{Performance comparison of TT decomposition ($\alpha=0.2$ rank=32) applied to different components and configurations. Single-layer defense is applied to the final normalization, attention output, and MLP layers, compared against multi-layer implementations on the last two and last five transformer block normalization layers.}
    \label{fig:layer-analysis}
\end{figure}

Figure \ref{fig:layer-analysis} illustrates the effectiveness of Tensor Train decomposition on various CLIP vision encoder components. The impact differs by component, with the final normalization layer offering the best defense in single layers. Extending to multiple layers improves results, with optimal outcomes from addressing the normalization layers of the last five blocks across all methods and datasets.



\subsubsection{Computational Efficiency}

\begin{table}[h]
\centering
\caption{Computational Efficiency of Different Tensor Decomposition Methods (Rank=64, $\alpha$=0.2)}
\label{tab:efficiency}
\setlength{\tabcolsep}{2pt}
\small
\begin{tabular}{lccccc}
\hline
\textbf{Method} 
  & \textbf{Rank} 
  & \textbf{Layers} 
  & \makecell{\textbf{Time/batch}\\\textbf{(ms)}} 
  & \textbf{Img/s} 
  & \textbf{Overhead} \\
\hline
\hline
No Defense & 0 & 0 & 87.07 & 367.51 & 1.00$\times$ \\
CP & 64 & 1 & 159.71 & 200.37 & 1.83$\times$ \\
Tucker & 64 & 1 & 240.17 & 133.24 & 2.76$\times$ \\
TT & 64 & 1 & 106.48 & 300.54 & 1.22$\times$ \\
CP Multi & 64 & 2 & 257.55 & 124.25 & 2.96$\times$ \\
TT Multi & 64 & 2 & 171.11 & 187.02 & 1.96$\times$ \\
TT Multi & 64 & 5 & 342.63 & 93.39 & 3.93$\times$ \\
\hline
\end{tabular}
\end{table}


Table \ref{tab:efficiency} shows the computational overhead of the decomposition methods, with single-layer TT achieving 1.22x overhead and 82\% throughput, outperforming Tucker and CP methods. This advantage extends to multilayer configurations, with 5-layer TT presenting a reasonable 3.93x overhead given its strong defensive capabilities, offering a tunable security-performance trade-off.


\section{Conclusion}
We presented a lightweight tensor decomposition defense that significantly improves VLM robustness against adversarial attacks without retraining or architectural modifications. Comprehensive experiments on COCO and Flickr30k using CLIP demonstrated effective adversarial filtering while preserving semantic content.

Key findings: (1) Tensor Train decomposition outperformed CP and Tucker methods, recovering up to 12.3\% performance with minimal computational overhead; (2) lower ranks (8-32) and residual strength ($\alpha=0.1-0.3$) provided optimal defense by filtering high-frequency adversarial components; (3) normalization layer targeting and multi-layer configurations (last five blocks) yielded strongest results; (4) TT decomposition achieved favorable efficiency with only 22\% overhead while maintaining 82\% throughput.

This plug-and-play approach offers practical advantages over methods requiring extensive retraining. Future work will explore adaptive rank selection and effectiveness in other VLM architectures and attack types.

\section{Limitations}
Although our tensor decomposition defense demonstrates promising results, several limitations must be acknowledged. First, our exploration was limited in time and resources, preventing exhaustive testing of all decomposition methods, rank values, and $\alpha$ parameters, and there might be better configurations outside our tested space. Second, we utilized three classical techniques (CP, Tucker, and TT), but advanced methods, such as Hierarchical Tucker or Block-Term Decomposition, could be more effective. Our tests focused on the CLIP model and PGD attacks; however, the effectiveness of this approach on other architectures or advanced attacks remains untested. Although retraining is not necessary, our approach adds inference-time computation that may hinder some real-time applications, especially with multi-layer configurations.

\section*{Acknowledgment}
Research was supported in part by the National Science Foundation under CAREER grant no. IIS 2046086, grant no. No. 2431569 and CREST Center for Multidisciplinary Research Excellence in CyberPhysical Infrastructure Systems (MECIS) grant no. 2112650. Research was also sponsored in part by the Army Research Office and was accomplished under Grant Number W911NF-24-1-0397. The views and conclusions contained in this document are those of the authors and should not be interpreted as representing the official policies, either expressed or implied, of the Army Research Office or the U.S. Government. The U.S. Government is authorized to reproduce and distribute reprints for Government purposes notwithstanding any copyright notation herein


\newpage

{
    \small
    \bibliographystyle{ieeenat_fullname}
    \bibliography{main}

\begin{thebibliography}{21}
\providecommand{\natexlab}[1]{#1}
\providecommand{\url}[1]{\texttt{#1}}
\expandafter\ifx\csname urlstyle\endcsname\relax
  \providecommand{\doi}[1]{doi: #1}\else
  \providecommand{\doi}{doi: \begingroup \urlstyle{rm}\Url}\fi

\bibitem[Bacciu and Mandic(2020)]{bacciu2020tensordecompositionsdeeplearning}
Davide Bacciu and Danilo~P. Mandic.
\newblock Tensor decompositions in deep learning, 2020.

\bibitem[Bulat et~al.(2021)Bulat, Kossaifi, Bhattacharya, Panagakis, Hospedales, Tzimiropoulos, Lane, and Pantic]{bulat2021defensivetensorization}
Adrian Bulat, Jean Kossaifi, Sourav Bhattacharya, Yannis Panagakis, Timothy Hospedales, Georgios Tzimiropoulos, Nicholas~D Lane, and Maja Pantic.
\newblock Defensive tensorization, 2021.

\bibitem[Comon(2009)]{comon2009tensordecompositionsstateart}
Pierre Comon.
\newblock Tensor decompositions, state of the art and applications, 2009.

\bibitem[Entezari and Papalexakis(2020)]{entezari2020tensorshieldtensorbaseddefenseadversarial}
Negin Entezari and Evangelos~E. Papalexakis.
\newblock Tensorshield: Tensor-based defense against adversarial attacks on images, 2020.

\bibitem[Harshman(1970)]{Harshman1970FoundationsOT}
Richard~A. Harshman.
\newblock Foundations of the parafac procedure: Models and conditions for an "explanatory" multi-model factor analysis, 1970.

\bibitem[Ji et~al.(2019)Ji, Wang, Li, and Liu]{8884203}
Yuwang Ji, Qiang Wang, Xuan Li, and Jie Liu.
\newblock A survey on tensor techniques and applications in machine learning.
\newblock \emph{IEEE Access}, 7:\penalty0 162950--162990, 2019.

\bibitem[Kossaifi et~al.(2019)Kossaifi, Panagakis, Anandkumar, and Pantic]{JMLR:v20:18-277}
Jean Kossaifi, Yannis Panagakis, Anima Anandkumar, and Maja Pantic.
\newblock Tensorly: Tensor learning in python.
\newblock \emph{Journal of Machine Learning Research}, 20\penalty0 (26):\penalty0 1--6, 2019.

\bibitem[Lahajal and S(2024)]{lahajal2024enhancingimageretrieval}
Naresh~Kumar Lahajal and Harini S.
\newblock Enhancing image retrieval : A comprehensive study on photo search using the clip mode, 2024.

\bibitem[Lorenz et~al.(2024)Lorenz, Harder, Strassel, Keuper, and Keuper]{lorenz2024detectingautoattackperturbationsfrequency}
Peter Lorenz, Paula Harder, Dominik Strassel, Margret Keuper, and Janis Keuper.
\newblock Detecting autoattack perturbations in the frequency domain, 2024.

\bibitem[Madry et~al.(2019)Madry, Makelov, Schmidt, Tsipras, and Vladu]{madry2019deeplearningmodelsresistant}
Aleksander Madry, Aleksandar Makelov, Ludwig Schmidt, Dimitris Tsipras, and Adrian Vladu.
\newblock Towards deep learning models resistant to adversarial attacks, 2019.

\bibitem[Mao et~al.(2023)Mao, Geng, Yang, Wang, and Vondrick]{mao2023understandingzeroshotadversarialrobustness}
Chengzhi Mao, Scott Geng, Junfeng Yang, Xin Wang, and Carl Vondrick.
\newblock Understanding zero-shot adversarial robustness for large-scale models, 2023.

\bibitem[Oseledets(2011)]{doi:10.1137/090752286}
I.~V. Oseledets.
\newblock Tensor-train decomposition.
\newblock \emph{SIAM Journal on Scientific Computing}, 33\penalty0 (5):\penalty0 2295--2317, 2011.

\bibitem[Papernot et~al.(2016)Papernot, McDaniel, Wu, Jha, and Swami]{papernot2016distillationdefenseadversarialperturbations}
Nicolas Papernot, Patrick McDaniel, Xi Wu, Somesh Jha, and Ananthram Swami.
\newblock Distillation as a defense to adversarial perturbations against deep neural networks, 2016.

\bibitem[Qian and Hu(2024)]{qian2024onlinezeroshotclassificationclip}
Qi Qian and Juhua Hu.
\newblock Online zero-shot classification with clip, 2024.

\bibitem[Radford et~al.(2021)Radford, Kim, Hallacy, Ramesh, Goh, Agarwal, Sastry, Askell, Mishkin, Clark, Krueger, and Sutskever]{radford2021learningtransferablevisualmodels}
Alec Radford, Jong~Wook Kim, Chris Hallacy, Aditya Ramesh, Gabriel Goh, Sandhini Agarwal, Girish Sastry, Amanda Askell, Pamela Mishkin, Jack Clark, Gretchen Krueger, and Ilya Sutskever.
\newblock Learning transferable visual models from natural language supervision, 2021.

\bibitem[Robey et~al.(2024)Robey, Wong, Hassani, and Pappas]{robey2024smoothllmdefendinglargelanguage}
Alexander Robey, Eric Wong, Hamed Hassani, and George~J. Pappas.
\newblock Smoothllm: Defending large language models against jailbreaking attacks, 2024.

\bibitem[Schlarmann et~al.(2024)Schlarmann, Singh, Croce, and Hein]{schlarmann2024robustclipunsupervisedadversarial}
Christian Schlarmann, Naman~Deep Singh, Francesco Croce, and Matthias Hein.
\newblock Robust clip: Unsupervised adversarial fine-tuning of vision embeddings for robust large vision-language models, 2024.

\bibitem[Tucker(1966)]{Tucker_1966}
Ledyard~R Tucker.
\newblock Some mathematical notes on three-mode factor analysis.
\newblock \emph{Psychometrika}, 31\penalty0 (3):\penalty0 279–311, 1966.

\bibitem[Wang et~al.(2025)Wang, Xie, Bartoldson, and Kailkhura]{wang2025doublevisualdefenseadversarial}
Zeyu Wang, Cihang Xie, Brian Bartoldson, and Bhavya Kailkhura.
\newblock Double visual defense: Adversarial pre-training and instruction tuning for improving vision-language model robustness, 2025.

\bibitem[Xie et~al.(2017)Xie, Wang, Zhang, Zhou, Xie, and Yuille]{xie2017adversarialexamplessemanticsegmentation}
Cihang Xie, Jianyu Wang, Zhishuai Zhang, Yuyin Zhou, Lingxi Xie, and Alan Yuille.
\newblock Adversarial examples for semantic segmentation and object detection, 2017.

\bibitem[Ye et~al.(2023)Ye, Kong, Yao, Ren, and Jiang]{ye2023videoquestionansweringusing}
Shuhong Ye, Weikai Kong, Chenglin Yao, Jianfeng Ren, and Xudong Jiang.
\newblock Video question answering using clip-guided visual-text attention, 2023.

\end{thebibliography}
}

\end{document}